# Harnessing AI for efficient analysis of complex policy documents: a case study of Executive Order 14110

**Mark A. Kramer, Allen Leavens[*], Alexander Scarlat**

MITRE Corporation, 202 Burlington Street, Bedford, MA

[*] Corresponding author

Email: aleavens@mitre.org

## Abstract

Policy documents, such as legislation, regulations, and executive orders, are crucial in shaping society. However, their length and complexity make interpretation and application challenging and time-consuming. Artificial intelligence (AI), particularly large language models (LLMs), has the potential to automate the process of analyzing these documents, improving accuracy and efficiency. This study aims to evaluate the potential of AI in streamlining policy analysis and to identify the strengths and limitations of current AI approaches. The research focuses on question answering and tasks involving content extraction from policy documents. A case study was conducted using Executive Order 14110 on "Safe, Secure, and Trustworthy Development and Use of Artificial Intelligence" as a test case. Four commercial AI systems were used to analyze the document and answer a set of representative policy questions. The performance of the AI systems was compared to manual analysis conducted by human experts. The study found that two AI systems, Gemini 1.5 Pro and Claude 3 Opus, demonstrated significant potential for supporting policy analysis, providing accurate and reliable information extraction from complex documents. They performed comparably to human analysts but with significantly higher efficiency. However, achieving reproducibility remains a challenge, necessitating further research and development.

## Introduction

Policy documents, including legislation, regulations, executive orders, operational guidelines, strategies, and action plans, serve a critical role in conveying requirements and direction for the audiences that are impacted by them. Interpreting and applying policy, proposed or enacted, can be difficult, time-consuming, and error-prone due to the length and complexity of policies. Reading, interpreting, and acting on policy documents can stretch or exceed human capacity. For example, it is estimated that there are 10,000 pages of "tiny regulatory type" regarding the implementation of the US Affordable Care Act.[1] As of April 2024, 16 states have enacted AI-related legislation, with more than 400 proposed AI-related bills across the US.[2] The relatively svelte EU AI Act[3] contains 90,000 words covering 459 pages.

Artificial intelligence (AI), especially large language models (LLMs), has the potential to automate the process of analyzing complex policy documents, reducing the time and effort required while improving accuracy and consistency. Policymakers and consumers may be able to quickly identify key points, extract relevant information, and generate summaries. This could enable more efficient decision-making, better policy implementation, and enhanced public understanding of important policy issues.

---

[1] https://www.washingtonpost.com/blogs/fact-checker/post/how-many-pages-of-regulations-for-obamacare/2013/05/14/61eec914-bcf9-11e2-9b09-1638acc3942e_blog.html
[2] https://www.cio.com/article/2081885/the-complex-patchwork-of-us-ai-regulation-has-already-arrived.html
[3] https://www.europarl.europa.eu/doceo/document/TA-9-2024-0138_EN.pdf

However, the application of AI to policy analysis is still an emerging field, and there are many questions and challenges that need to be addressed, among them: how accurate and reliable are LLM-based AI systems[4] in extracting information from complex policy documents? Can they handle the nuances and ambiguities of legal and policy language? How do they compare to human analysts in terms of performance and efficiency?

To address these questions, we conducted a case study using Executive Order 14110 on "Safe, Secure, and Trustworthy Development and Use of Artificial Intelligence." This Executive Order, issued by the President of the United States in October 2023, spans over 20,000 words, and contains a wide range of directives and initiatives related to AI development and governance. We used four commercial, state-of-the-art AI systems to analyze the document and answer a set of representative policy questions and perform typical policy analysis tasks. We then compared the performance of the AI systems to manual analysis conducted by human experts, evaluating metrics such precision, recall, and F1 score. In addition, we analyzed the AI systems' responses to identify strengths, weaknesses, and potential areas for improvement.

Our research aimed to provide insights into the potential of AI for streamlining policy analysis and limitations of current AI approaches. We focused only on question answering and tasks involving content extraction — not summarization, interpretation, impact, or other analyses — with the goal of determining if current AI systems can achieve comparable or better performance than human analysts, while significantly reducing the time and effort required. We did not investigate the potential of teaming between human analysts and AI systems.

## Prior Work

The application of artificial intelligence (AI) to policy analysis is an emerging field with the potential to revolutionize the way policies are formulated, implemented, and evaluated. Rapid advancements in AI techniques, such as natural language processing (NLP), machine learning, and knowledge extraction, offer a promising solution to streamline the analysis of complex policy documents (Safaei and Longo, 2024). Government agencies have been testing out use of AI tools to support these types of efforts, such as highlighted by examples in a recent report for the Administrative Conference of the United States (Sharkey, 2023).

Researchers have investigated the use of automated text analysis for enabling systematic assessment of political science content collections, such as the text of debates and discussion of legislation (Grimmer & Stewart, 2013). Branting et. al. (2019) assessed the predictive accuracy of an automated approach to classification of directive sentences in policy documents. Lopez. et. al. (2021) assessed AI use and knowledge extraction from policy in efforts to improve identification of fraud, waste, and abuse in healthcare. Cook et. al. (2024) examined use of AI within a transdisciplinary approach to analyze expressions in regulatory texts.

Similarly, AI tools are being used to analyze legal documents, predicting case outcomes, identifying relevant precedents, and automating legal research tasks (Aletras et al., 2016, Rodgers et. al. 2023). Saha et. al. (2017) investigated the potential for automated knowledge extraction from lengthy legal documents, using the Federal Acquisition Regulations as a test case. Ioannidis et. al. (2023) tested a specialized approach to using generative AI in legal and regulatory compliance.

Going beyond policy text analysis and development, some investigations have focused more on the utility of AI for modeling and forecasting policy outcomes. Researchers have employed machine learning techniques such as regression analysis, decision trees, and neural networks to predict the impacts of policy changes on various economic and social indicators (Kleinberg et al., 2015). Additionally, AI-powered simulations and agent-based models allow for testing policy scenarios and exploring potential unintended consequences before implementation. For example, researchers have used AI to model the impact of tax policies on income inequality and economic growth, providing valuable insights for policymakers (Mullainathan & Spiess, 2017). AI techniques, particularly causal inference methods like causal forests and double machine learning, are increasingly used to address challenges in policy evaluation, such as selection bias and confounding variables (Athey & Imbens, 2017).

---

[4] We use the term "AI system" rather than "AI model" because in general, commercial products have pre- and post-processing steps rather than a purely stand-alone LLM.

The use of AI in policy analysis also presents challenges, including concerns about data bias, algorithmic transparency, and the potential displacement of human expertise. LLMs and the applications that use them may be useful for more efficient evaluation of policies and potentially could help identify bias impacting certain demographic groups covered by a certain policy; however, AI systems also can lack sufficient capability to judge nuances in policy language compared to humans, may introduce data privacy concerns, and could lead to threats based on biases and limitations built into the AI models themselves (Morita et. al., 2023).

# Methodology

To investigate the potential of AI for analyzing complex policy documents, we conducted a case study using Executive Order 14110 ("EO") on "Safe, Secure, and Trustworthy Development and Use of Artificial Intelligence" as our primary data source. This executive order, issued by the President of the United States on October 30, 2023, represents a comprehensive and multifaceted policy document that covers a wide range of topics related to AI development, governance, and use. The EO consists of 13 sections, spanning approximately 20,000 words.

For our analysis, we tested four AI systems that were state-or-the-art as of April 2024: Claude 3 Opus ("Claude") by Anthropic, ChatGPT-4 ("GPT4") by OpenAI, Gemini Pro 1.5 ("Gemini") by Google, and Command R+ by Cohere ("Cohere"). The first three are top-performing foundation-model systems and the fourth represents an implementation of retrieval augmented generation (RAG). RAG is a method of dealing with large text corpora and/or limited context windows by using an external information repository (relational database, vector database, knowledge graph, etc.) and retrieving relevant passages based on a given query, prior to activating the LLM. The EO is short enough to fit into the context windows of Claude (200,000 tokens), GPT4 (undisclosed, estimated 32,000 tokens), and Gemini (64,000 tokens) without truncation or resorting to retrieval augmentation.

Preliminary tests were conducted in April 2024 in which we queried the AI systems about the EO without uploading the document or providing it in the prompt. Claude, with a knowledge cutoff in August 2023, and Gemini, with a cutoff in November 2023, both explicitly stated they had no knowledge of the EO and were unable to summarize the document. Cohere, with a knowledge cutoff in January 2023, likewise demonstrated no knowledge of the EO. However, Cohere failed to recognize its lack of knowledge and, when prompted summarize the EO, created a false narrative dealing with telecommunications and supply chains. GPT4, with a knowledge cutoff in December 2023, was able to summarize and answer questions about the EO without being provided with the document. Interestingly, being trained on the EO did not prove to be an advantage for GPT4, as it subsequently underperformed compared to the other AI systems.

To evaluate the performance of these AI systems in analyzing the EO, two tests were devised:

- A set of 100 multiple-choice questions designed to assess the systems' ability to extract specific information, identify key points, and synthesize complex ideas (see Supplement). Candidate questions and answers were initially generated by Claude, followed by manual editing for accuracy and clarity. Because of Claude's involvement in creating the questions, Claude was excluded from the set of AI systems evaluated on these questions.
- A set of six tasks requiring analysis of the entire EO and extraction of lists of items (see Supplemental Material). These tasks were inspired by prior work performed by MITRE on behalf of a government organization, which required multiple people spending tens of hours poring over the EO and its companion document, draft memorandum M-24-10 by the Office of Management and Budget (OMB). The time and effort expended in that effort triggered the authors' interest in more efficient methodologies.

To establish ground truth on these tests, the authors independently analyzed the EO, compared answers, and resolved discrepancies. We also rechecked the ground truth when the consensus of AI systems suggested different answers. The resulting answers are given in the Supplement. With ground truth established, answers could be graded without further subjective judgment.

## Multiple-Choice Questions

For each AI system, we uploaded a PDF version of Executive Order 14110, obtained from a government web site.[5] We then posed the questions using the chat interface of each model. For the multiple-choice question set, all 100 questions were introduced simultaneously[6] in tabular format, either MS Excel or MS Word, depending on the capabilities of the AI system. The following prompt was used[7]:

> *I have uploaded a set of 100 questions about Executive Order 14110. Please answer each question referencing the uploaded Executive Order document, selecting the best single answer to each question. Return the result as a table, with the question number in column 1 and the answer as a letter (A, B, C, D, or E) in column 2. No other output is necessary. Answer all the questions in order.*

All the AI systems were able to follow these directions and return a table containing 100 answers. The performance on the multiple-choice questions were measured in terms of percent correct.

Because LLM text generation algorithms involve probabilistic determination of the next output word, identical inputs can lead to variations in outputs. To measure replicability, we ran the question set several times, using a new chat session each time to eliminate conversational memory effects. Both the variation in percent correct over the entire question set and the run-to-run variation in answers to each question were measured.

## Information Extraction Tasks

The six retrieval tasks were run one at a time. We chose to use the chat interfaces of these systems because of the likelihood that policy analysts would use the chat modality, rather than an application programming interface. A new chat was initiated for each task. The tasks are as follows:

1. List all Federal Government Organizations (Departments, Agencies, Administrations, Centers, Institutes, Services, and Offices) mentioned in Executive Order 14110 (for example, National Institute of Standards and Technology (NIST), Department of Homeland Security, Office of Management and Budget). Include the organization if individual affiliated with the organization is mentioned (for example, Secretary, Director, Chief AI Officer). Do not include boards, councils, groups, programs, or committees.

2. List all actions and responsibilities in Executive Order 14110 that are explicitly assigned to the Department of Health and Human Services (HHS) or its subcomponents (such as CDC, CMS, and FDA). Include actions assigned to the Secretary of HHS or other HHS personnel. Give a fully-numbered reference for each identified action (for example, 4.1, 6.b.i, 4.2.a.i.B, or 5.3.b.iii.D).

3. Create a list of actions in Executive Order 14110 that must be completed within 6 months of the issuance of the order. Output the result as a three-column table giving (1) the item number (for example, 4.5.a, 7.2.b.ii, 8.b.iii, etc.), (2) the specified deadline (e.g., 45 days, 60 days, etc.), and (3) the action to be performed.

4. Which specific actions in Executive Order 14110 deal with assuring AI is applied equitably, without bias, in a non-discriminatory manner? Return a list of actions, a fully-numbered reference for each identified action (for example, 4.1, 6.b.i, 4.2.a.i.B, or 5.3.b.iii.D), and the name of the agency involved with implementing the measure.

---

[5] https://www.federalregister.gov/documents/2023/11/01/2023-24283/safe-secure-and-trustworthy-development-and-use-of-artificial-intelligence

[6] It is possible that performance may be affected by introducing all the questions at once, versus introducing the questions individually or in smaller groups. This was not investigated but may be worthy of further study.

[7] The way an LLM is prompted can have a substantial impact on LLM performance, however, prompt engineering was not explored in the current study.

5. Identify actions in Executive Order 14110 that apply to all US Federal Government agencies. Return a list of actions, a fully-numbered reference for each identified action (for example, 4.1, 6.b.i, 4.2.a.i.B, or 5.3.b.iii.D), and the name of the agency involved with implementing the measure.

6. List all new reports, guidance, regulations, or similar documents that are to be developed in response to Executive Order 14110. Return a list with a fully-numbered reference for each identified document (for example, 4.1, 6.b.i, 4.2.a.i.B, or 5.3.b.iii.D).

The prompts and correct answers for each task are given in the Supplement.

The answers to the retrieval tasks were lists that could be compared to the ground truth in terms of true positives (items in the answer also in the ground truth), false positives (items in the answer not in the ground truth), and false negatives (items in the ground truth not in the answer). There were converted the standard measures precision, recall, and F1 score (Van Rijsbergen, 1979).

To establish a baseline of human performance, we conducted human trials with three volunteers using the first three of the information extraction tasks. Volunteers were given 4 hours to complete the tasks, which everyone used in full. These individuals all had prior experience with health policy analysis and/or planning and were given the same information as the AI systems. The human analysts were instructed that they may consult reference sources on the web when composing the responses, but to avoid the use of any AI-based tools.

# Results

## Multiple-Choice Questions

For the 100 multiple-choice questions, we compared GPT4, Cohere, and Gemini, sidelining Claude since it was the original source of these questions. We conducted five runs with each model, as summarized in Table 1.

Of those tested, Gemini proved to be the most accurate model by a substantial margin, with approximately twice the percentage of correct answers than GPT4 or Cohere over the five runs (71.4% versus 38.4% and 37.6% respectively). Gemini's accuracy varied from run to run by as much as 24% and GPT's by 33%, whereas the Cohere's accuracy varied by at most 4%. This may be a result of the RAG approach reducing the context provided to the LLM through semantic search of the EO, yielding the same EO paragraphs on each run, and simultaneously reducing the amount of information provided to the LLM to only those relevant passages.

In addition to looking at overall accuracy run-to-run, we also looked at replicability, namely, how answers to a single question varied run-to-run. We used a metric for replicability ranging linearly from 0% (five different answers) to 100% (five answers in agreement).[8] GPT4 was the least replicable (41.0%, or 2.64 answers agreeing across 5 runs), while Cohere was the most replicable (81.5% or 4.26 answers agreeing, on average), closely followed by Gemini (80.0% replicability).

The consensus solution is the answer most often produced across multiple runs. Consensus does not exist when there are five different answers across the five runs. Looking at each model separately, the consensus solution was marginally beneficial or counterproductive. GPT4 averaged 38.4% accuracy in individual trials versus 41.0% accuracy using the consensus of five trials. Gemini averaged 71.4% accuracy per run, but only 66.0% using the consensus of the five runs. Pooling all 15 runs together over all three AI systems, the consensus was 65.0% accurate, lower than the average accuracy of Gemini (71.4%) but higher than GPT4 or Cohere.

Lastly, we found that consistent answering across multiple runs correlated strongly with the likelihood of a correct answer for Gemini and GPT4. With Gemini, when 3 or fewer answers agreed, the chance that the consensus answer was correct was

---

[8] Conversion to replicability percentage from the number of answers agreeing ($n$) and the number of question choices ($q$) is $(n-1)/(q-1)$ since 0% replicability (all different answers) corresponds to $n=1$ (not $n=0$).

20.6%, but when 4 of 5 answers agreed, the consensus answer was correct 67% of the time, and when 5 out of 5 answers agreed, the consensus answer was correct was 92.9% of the time. For GPT, when 2 answers agreed, there was only a 13.6% chance that the consensus answer was correct, when 3 answers agreed, this jumped to 60%, and when 4 or 5 answers agreed, there was an 84.6% chance the consensus answer was correct. For Cohere, this trend was much weaker, with 32.3% correct when 2-3 answers agreed, 50% correct when 4 answers agreed, and 42.6% correct when all 5 answers agreed.

*Table 1. Results of Multiple-Choice Test*

| Method | Range of Accuracy over 5 runs (%) | Average Accuracy over 5 runs (%) | Average per-answer replicability (%) | Consensus answer accuracy (%) | Correlation between replicability and the consensus answer accuracy |
|---|---|---|---|---|---|
| GPT4 | 22.0 to 55.0 | 38.4 | 41.0 | 41.0 | 0.549 |
| Cohere | 36.0 to 40.0 | 37.6 | **81.5** | 40.0 | 0.095 |
| Gemini | 62.0 to 86.0 | **71.4** | 80.0 | **66.0** | 0.664 |
| Pooled | n/a | 65.0 | 55.9 | 65.0 | 0.352 |

## Information Extraction Tasks

The six retrieval tasks were run with all four AI systems. Human analysts, all MITRE employees, addressed only the first three of these tasks, due to limited time availability. One run with each AI system was conducted. Results are summarized in Table 2.

On the first three tasks, where a human baseline was available, the performance of humans, Gemini, and Claude were comparable with F1 scores of 0.83 versus 0.82 and 0.82, respectively. Based on F1 scores, Gemini outperformed humans on Tasks 1 and 2, while humans outperformed Gemini by a significant margin on Task 3.

Over all six tasks, Gemini outperformed the others, with an average F1-score across of 0.74, followed by Claude at 0.68. GPT4 was the poorest performer by a large margin, which is somewhat surprising, since it has been regarded as the premier LLM model. Cohere ranked between Claude and GPT4.

*Table 2. Results for Retrieval Tasks (P=precision, R=recall, F= F1-score)*

|  | Task 1 | | | Task 2 | | | Task 3 | | | Average (Tasks 1-3) | | |
|---|---|---|---|---|---|---|---|---|---|---|---|---|
|  | P | R | F | P | R | F | P | R | F | P | R | F |
| GPT4 | 0.79 | 0.26 | 0.39 | 0.75 | 0.33 | 0.46 | 0.29 | 0.07 | 0.11 | 0.61 | 0.22 | 0.32 |
| Cohere | 0.71 | 0.70 | 0.71 | 0.78 | 0.78 | 0.78 | 0.66 | **0.83** | 0.73 | 0.72 | **0.77** | 0.74 |
| Claude | **1.00** | 0.77 | 0.87 | 0.88 | 0.78 | 0.82 | 0.85 | 0.69 | 0.76 | 0.91 | 0.75 | 0.82 |
| Gemini | 0.95 | **0.85** | **0.90** | **1.00** | **1.00** | **1.00** | 0.92 | 0.4 | 0.55 | **0.96** | 0.75 | 0.82 |
| Humans | 0.87 | 0.81 | 0.84 | 0.86 | 0.78 | 0.81 | **1.00** | 0.73 | **0.84** | 0.91 | **0.77** | **0.83** |

|  | Task 4 | | | Task 5 | | | Task 6 | | | Average (Tasks 1-6) | | |
|---|---|---|---|---|---|---|---|---|---|---|---|---|
|  | P | R | F | P | R | F | P | R | F | P | R | F |
| GPT4 | 0.71 | 0.23 | 0.34 | 0.44 | 0.07 | 0.12 | 0.29 | 0.25 | 0.27 | 0.55 | 0.20 | 0.28 |
| Cohere | 0.65 | 0.50 | 0.56 | 0.40 | 0.38 | 0.39 | 0.15 | 0.50 | 0.23 | 0.56 | 0.62 | 0.57 |
| Claude | 0.77 | 0.45 | 0.57 | **0.93** | 0.46 | 0.61 | 0.36 | **0.63** | 0.45 | 0.80 | 0.63 | 0.68 |
| Gemini | **0.93** | **0.59** | **0.72** | 0.92 | **0.58** | **0.71** | **0.57** | 0.50 | **0.53** | **0.88** | **0.65** | **0.74** |

# Discussion

## Multiple-Choice Answering

In the multiple-choice questions, we noticed patterns in what distinguished "easy" questions from "hard" ones across all LMMs. In general, the easy questions tended to be about:

- High-level principles, priorities, and goals emphasized throughout the document.
- General requirements, actions, or responsibilities directed at major entities and agencies, often stated clearly and repeatedly.
- Key definitions and concepts that are explicitly provided in the order.

On the other hand, hard questions more often related to:

- Specific technical details, examples, or considerations that are only briefly mentioned or buried deeper in the full text.
- Actions, requirements, and responsibilities assigned to lesser-known entities, sub-agencies, or smaller provisions of the order.
- Technical specifics of how certain initiatives or requirements will be implemented.
- Directives or responsibilities that are phrased in a less explicit or direct way.

Cohere departed somewhat from these generalities, answering some detailed questions but missing some clearly stated points, suggesting that the ability to find key passages during the semantic search phase was influencing outcomes.

Run-to-run, Cohere provided the most consistent answers, but not the most accurate. By their nature, RAG methods depend on a deterministic search process to find a limited number of relevant passages, which are then introduced to the LLM. Metaphorically, instead of the LLM searching for a needle in a haystack, it only needs to search through a bale of hay extracted by from the haystack by the semantic search. If the hay bale contains the needle, the LLM has a high likelihood of finding it, but if the needle is not there, the LLM will never find it.

Using consensus over multiple runs to determine the answer to a question provided mixed results, with either a marginal increase in accuracy or a degradation compared to the average accuracy over the same runs. There was, however, a strong correlation between run-to-run agreement and the accuracy of the answer (except for Cohere). Put another way, run-to-run disagreement flags potentially inaccurate answers.

## Information Extraction Tasks

Gemini and Claude demonstrated the most comprehensive understanding of the EO, consistently providing concise, accurate, and detailed responses across all questions, using formatting like tables and bullet points to enhance readability. For example, Claude addressed Task 1 (listing government organizations mentioned in the EO by name or indirectly through individuals belonging to the organization) by listing each organization, the individuals mentioned, and suborganizations:

*The following Federal Government Organizations are mentioned in Executive Order 14110:*
1. *Department of Energy*
   - *Secretary of Energy*
2. *Department of Homeland Security*
   - *Secretary of Homeland Security*
   - *Director of the Cybersecurity and Infrastructure Security Agency*
3. *Department of the Treasury*
   - *Secretary of the Treasury*
4. *Department of Defense*
   - *Secretary of Defense*

5. *Department of Justice*
   - *Attorney General*
   - *Assistant Attorney General in charge of the Civil Rights Division*
6. *Department of Commerce*
   - *Secretary of Commerce*
   - *Under Secretary of Commerce for Intellectual Property and Director of the United States Patent and Trademark Office*
   - *National Institute of Standards and Technology (NIST)*
     - *Director of NIST*
   - *National Telecommunications and Information Administration*
     - *Assistant Secretary of Commerce for Communications and Information*

*(et cetera)*

Cohere provided detailed answers but occasionally included irrelevant information or misinterpreted the scope of the question. For example, on Task 3 (actions required within 180 days), Cohere included actions whose deadlines extended beyond 180 days. On Task 5 (actions to be taken by all agencies), Cohere, Claude, and Gemini mistakenly included topics from Section 10, which is specifically directed to OMB, because there is language directing OMB to develop guidance applicable to all agencies. GPT4 provided fluent summaries but at times seemed to stop short of analyzing the entire document. For example, on Task 5, GPT4 provided answers exclusively from Section 4 of the EO, although answers were contained in multiple other sections.

## Other Observations

All four AI systems tested were able to upload and work with the EO in the original PDF format. All systems were able to deal with the complex numbering system encountered in the EO, where paragraphs and subparagraphs are identified using a mixture of letters, numbers, and roman numerals, up to five levels deep, for example, 10.1(b)(viii)(C). Only the lowest-level identifier is in direct proximity to the paragraph it identifies. Determining the full paragraph reference involves backtracking through multiple levels of the document hierarchy until reaching a top-level heading (see Figure 1). Manually backtracking is tedious and error prone, especially when the top-level heading occurs several pages previously. Despite this apparent difficulty, the AI systems provided full and accurate document references in almost all cases.

Safety filtering should have played no role in this case study, but at the beginning of several chat sessions, Gemini would halt based on "unsafe sexually explicit material." It goes without saying that the EO does not contain sexually explicit material. However, it does refer to preventing AI from generating child sexual abuse material. Apparently, these words were sufficient to trigger Gemini's safety filters. Fortunately, Gemini allows filtering sensitivity to be lowered, so the requisite tasks could be carried out.

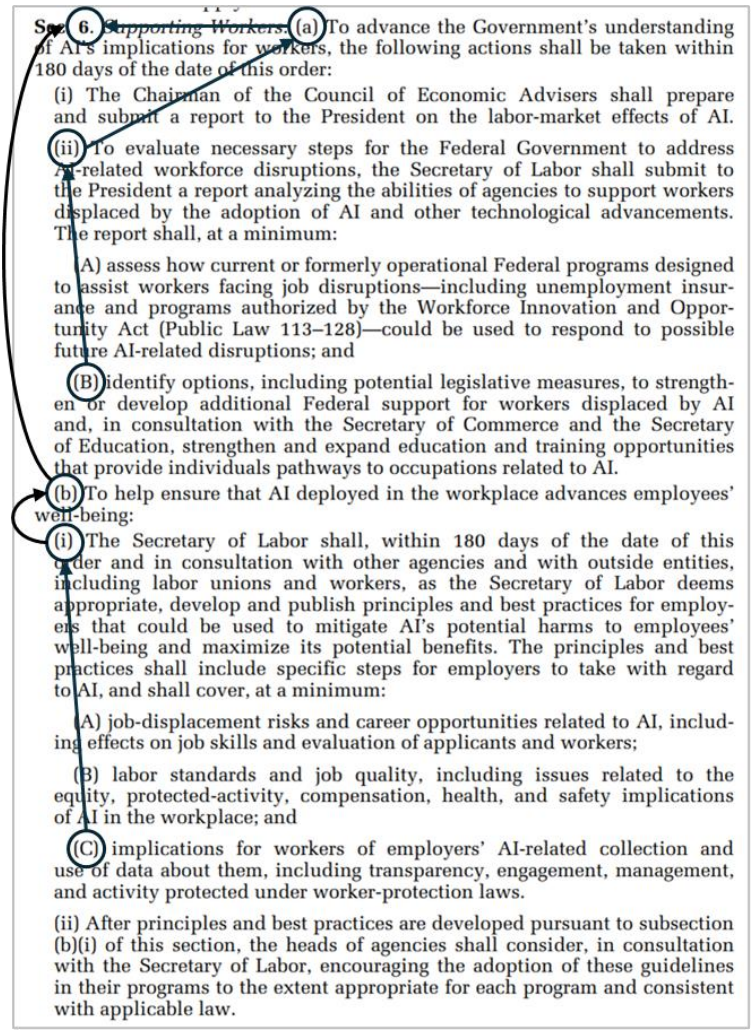

*Figure 1. Screenshot of the Executive Order PDF with two examples of backtracking to determine full paragraph references*

# Limitations

While this study provides valuable insights into the potential of AI in policy analysis, there are several limitations to consider:

- There are several aspects that remain to be investigated. For example, the study did not attempt to improve accuracy or replicability through prompt engineering. We did not explore the effect of introducing the multiple-choice questions individually or in smaller batches. In the information retrieval tasks, we did not investigate the effect if retrieval could be improved by dividing the EO to smaller subsections, to limit the search range. The study did not explore the effect of hyperparameters, such as temperature, top-k, and top-p, which could influence the results.
- The research was limited to a single case study, Executive Order 14110. While this document is complex and representative of many policy documents, the findings may not generalize to all types of policy documents, particularly those with different structures, language styles, or subject matters.
- Larger, multiple-document corpora, particularly those that exceed current context window sizes, would provide a different test of AI systems' capabilities and limitations.

- The study focused only on question answering and tasks involving content extraction from policy documents. Other important aspects of policy analysis, such as summarization, interpretation, impact analysis, and other types of analyses, were not considered. Therefore, the full potential of AI in policy analysis has not been fully captured.
- The study did not investigate the potential of teaming between human analysts and AI systems, which could potentially lead to better results than either could achieve alone.
- Only four commercial AI systems were evaluated. There are many other AI systems, including open-source alternatives, that may perform differently. It is a snapshot of one point in time (March-April 2024) in a rapidly-evolving field.

# Conclusion

This study presented a small case study on the accuracy and reliability of LLM-based AI systems used to extract information from complex policy documents and compared their performance to that of human analysts. The results show that leading foundation models can operate with levels of retrieval and precision commensurate with human performance. Although other aspects of policy analysis such as summarization, interpretation, and impact analysis were not investigated, this analysis suggests that at least some types of policy analysis by LLM-based AI systems may be viable. However, since there are significant variations between AI systems, and run-to-run variability in the answers obtained from the same AI system, the most difficult challenge will be advancing technically to the point where a human policy analyst can have confidence using these tools.

At present, Gemini and Claude appear to be the most capable models for accurate and reliable information extraction from complex documents like Executive Order 14110. They share the ability to understand the context, follow instructions, and provide well-structured responses. Gemini demonstrated retrieval and precision commensurate with human levels of performance, but much faster, accomplishing tasks that took human reviewers 4 hours in a few minutes.

While Cohere shows potential, it was not able to achieve the same level of accuracy. However, as a RAG architecture, it may be the only system among the ones tested capable of scaling to very large text corpora. GPT4, in its current state, appears less suitable for policy analysis tasks demanding precision and faithfulness to source material.

Our testing on 100 multiple-choice questions showed significant run-to-run variability. Achieving acceptable levels of reproducibility and trustworthiness remains a critical challenge that necessitates further research and development. Techniques such as self-consistency (Wang et al., 2023, Chen et al., 2023), self-evaluation (Ren et al, 2023) have shown promise in reducing variability.

Further research could involve testing other AI models, including open-source alternatives, mixture-of-experts systems, and other emerging architectures. Expanding the scope to include additional tasks such as summarization, consistency checking, and question answering on larger corpora using augmented LLMs would provide a more comprehensive evaluation of these systems' capabilities and limitations.

# Supplemental Materials

## Multiple-Choice Questions

The following prompt was used with the following question set: *I have uploaded a set of 100 questions about Executive Order 14110. Please answer each question referencing the uploaded Executive Order document, selecting the best single answer to each question. Return the result as a table, with the question number in column 1 and the answer as a letter (A, B, C, D, or E) in column 2. No other output is necessary. Answer all the questions in order.*

| | Question | Answer A | Answer B | Answer C | Answer D | Answer E | Answer | Source |
|---|---|---|---|---|---|---|---|---|
| 1 | Which of the following is not a potential benefit of responsible AI use mentioned in the document? | Making the world more secure | Promotion of social justice | Making the world more prosperous | Making the world more innovative | Solving urgent challenges | B | 1, 2(b), 2(d) |
| 2 | Under this order, what type of testing should AI systems undergo to ensure safety and security? | Randomized controlled trials | Post-deployment performance monitoring | Running standard benchmarks | White-box testing | All of the above | B | 2(a), 4.1(a)(i), 4.2(a)(i)(C) |
| 3 | Which of these is not listed as an example of how AI could be misused in the workplace? | Encouraging undue worker surveillance | Stifling market competition | Enhancing job quality | Causing harmful labor disruptions | Introducing new health and safety risks | C | 2(c) |
| 4 | The order directs agencies to use policy and technical tools for what purpose related to privacy? | To enable data sharing with third parties | To facilitate AI-assisted surveillance | To protect against improper data collection and use | To streamline data access for AI researchers | To mandate use of specific privacy software | C | 2(f), 9(a) |
| 5 | What does the order require of companies developing potential dual-use foundation models? | To make their model code open source | To provide ongoing information to the government | To obtain a license before beginning development | To pay a special tax on model training costs | To use only government-approved training data | B | 4.2(a)(i), 4.2(b) |
| 6 | Which of the following best describes a key aim of the AI and Technology Talent Task Force? | To increase AI education in K-12 schools | To establish government-funded AI PhD programs | To create an AI research grants database | To accelerate hiring of AI talent in the government | To develop standardized AI job application forms | D | 10.2(b) |
| 7 | According to the document, what should employers do to mitigate AI's potential harms to employees? | Provide AI ethics training to all staff | Require employees to sign AI liability waivers | Ban the use of AI for employee monitoring | Establish principles and best practices on job displacement | Replace human resources staff with AI systems | D | 6(b)(i), 6(b)(ii) |
| 8 | The order encourages agencies with regulatory authority over critical infrastructure to take what action? | Require use of open-source AI systems only | Mandate use of agency-developed AI software | Evaluate potential AI risks and mitigate vulnerabilities | Ban AI use in critical infrastructure | Exempt critical infrastructure from AI regulation | C | 4.3(a)(iii), 4.3(a)(iv) |
| 9 | Which of these is mentioned as a potential risk of foundation models with widely available weights? | Economic harms to AI startups | Making models available to adversaries | Removal of model safeguards | Challenge of assigning legal liability | Acceleration of socially beneficial AI research | C | 4.6(a), 4.6(b) |
| 10 | What is a key focus area for OMB's future guidance on government AI use directed by the order? | Standardizing AI procurement contract language | Setting minimum accuracy rates for AI systems | Issue guidance on AI uses that impact rights or safety | Requiring open-source publication of AI code | Mandating third-party auditing for all AI systems | C | 10.1(a), 10.1(b)(iii) |
| 11 | Which of the following is not listed as a potential harm from irresponsible AI use? | Displacement of workers | Stifling of competition | Reduction of economic inequality | Exacerbation of disinformation | Posing national security risks | C | 1 |
| 12 | The order states that effective AI leadership involves pioneering what? | Technological advancements | Military AI applications | Systems and safeguards for responsible deployment | AI ethics principles | International AI regulations | C | 2(h) |
| 13 | Which of these is included in the definition of an AI system? | Software | Hardware | Data systems | Utilities using AI | All of the above | E | 3(e) |

| # | Question | A | B | C | D | E | Answer | Ref |
|---|---|---|---|---|---|---|---|---|
| 14 | What type of AI systems are presumed to impact rights or safety under the order's guidance? | Systems used for entertainment | Systems used for scientific research | Systems used for employee productivity tracking | Systems used for weather forecasting | This is to be determined by OMB | E | 10.1(b)(v) |
| 15 | The AI Bill of Rights was developed by which entity? | Department of Justice | Federal Trade Commission | Department of Labor | Consumer Financial Protection Bureau | None of the above | E | Not mentioned in the order |
| 16 | Which of these best describes "privacy enhancing technology"? | Software that prevents targeted advertising | Hardware for secure data storage | Techniques to protect privacy in data processing | Algorithms for faster data processing | Blockchain-based anonymous transactions | C | 3(z) |
| 17 | As a risk management practice, the order requires agencies to conduct what type or types of assessment for AI systems? | Data quality | Economic impact | Algorithmic discrimination | All of the above | Only A and C | E | 10.1(b)(iv) |
| 18 | What is the primary purpose of the Artificial Intelligence Safety and Security Board? | To approve AI systems before deployment | To enforce penalties for AI misuse | To provide advice on AI security in critical infrastructure | To fund AI security research | To develop government AI use guidelines | C | 4.3(a)(v) |
| 19 | Which entity is directed to establish guidelines for AI red-teaming? | National Institutes of Health | National Institute of Standards and Technology | National Science Foundation | Department of Homeland Security | Federal Bureau of Investigation | B | 4.1(a)(ii) |
| 20 | The order directs the creation of what, related to biosecurity and AI? | Database of AI-assisted biosecurity incidents | Guidelines for AI use in biodefense research | Testbed for evaluating AI biotechnology tools | Study on AI and biosecurity risks and opportunities | Interagency task force on AI and biosafety | D | 4.4(a)(ii) |
| 21 | What type of content are agencies discouraged from banning under this order? | Social media posts | Online advertisements | Generative AI outputs | Violent video games | Cryptocurrency transactions | C | 10.1(f)(i) |
| 22 | Which agency is tasked with developing an AI assurance framework for healthcare? | Centers for Disease Control and Prevention | Food and Drug Administration | National Institutes of Health | Department of Veterans Affairs | Department of Health and Human Services | E | 8(b)(ii) |
| 23 | The order directs the Department of Commerce to engage with which entities on global technical standards for AI outside of military and intelligence use? | State and local governments | Key international allies | Standards development organizations | All of the above | Only B and C | E | 11(b) |
| 24 | Which of these is a not priority area for government use of AI under the order? | Civil rights and discriminiation | Labor-market effects of AI | Military and defense applications of AI | Chemical, Biological, Radiological, Nuclear (CBRN) threats | Critical infrastructure and cybersecurity | C | Not mentioned as a priority in the order |
| 25 | The order establishes what type of body within the Executive Office of the President? | National AI Research Coordination Council | White House Artificial Intelligence Council | President's Council on Science and Technology | Interagency AI Ethics Committee | None of the above | B | 12(a) |
| 26 | According to the order, what should be a focus of government AI safety program reporting? | Number of AI systems deployed | Amount of funding allocated to AI programs | Incidents of harm or discrimination caused by AI | Demographic diversity of AI system users | Comparison to AI adoption by other nations | C | 8(b)(iv)(A) |

| # | Question | A | B | C | D | E | Answer | Ref |
|---|----------|---|---|---|---|---|--------|-----|
| 27 | The order directs the development of an "AI in Global Development Playbook" by whom? | Department of Homeland Security | US Agency for International Development | Department of Defense | Department of Energy | Office of the US Trade Representative | B | 11(c)(i) |
| 28 | Which of these is included in information companies may need to report relating to dual-use AI? | Planned activities related to AI foundation model training | Resumes of AI researchers employed | Financial investment sources | Nationalities of key technical staff | Data sources used for training models | A | 4.2(a)(i)(A) |
| 29 | The order encourages agencies to enable staff access to generative AI for what purpose? | Analyzing public sentiment on social media | Generating reports and official correspondence | Automating software development processes | Experimentation and routine low-risk tasks | Optimizing resource allocation and budgeting | D | 10.1(f)(i) |
| 30 | Which agency is directed to develop a framework for AI security risk assessment? | National Security Agency | Office of the Director of National Intelligence | Department of Homeland Security | Federal Bureau of Investigation | Central Intelligence Agency | C | 4.4(b)(iv)(A) |
| 31 | The order directs the establishment of what type of program related to AI in biomedicine? | AI-assisted vaccine development initiative | Biomedical AI startup incubator | AI biosecurity research funding program | AI bioethics advisory committee | AI safety monitoring program for healthcare entities | E | 8(b)(iv) |
| 32 | Which of these is not mentioned as an example of AI's potential to improve climate and energy resilience? | Improve community preparedness | Optimizing renewable energy deployment | Reducing permitting delays | Improving the efficiency of fossil fuel extraction | Enhancing the resilience of the electric grid | D | 5.2(g) |
| 33 | The order requires agencies to appoint what type of official? | Chief AI Ethics Officer | Chief AI Research and Development Officer | Chief AI Safety and Security Officer | Chief AI Talent and Workforce Development Officer | Chief Artificial Intelligence Officer | E | 10.1(b)(i) |
| 34 | Which entity is encouraged to consider using AI to combat robocalls? | National Institutes of Health | Federal Trade Commission's Bureau of Consumer Protection | Federal Communications Commission | National Institute of Standards and Technology | Environmental Protection Agency | C | 8(a), 8(e)(iv) |
| 35 | The order establishes conditions for reportable AI systems involving what metrics? | Number of parameters in the model | Amount of computing power used for training | Both number of parameters and computing power | Number of developers on the project team | Expected societal impact of the model | B | 4.2(b) |
| 36 | What is the stated purpose of the National AI Research Resource Task Force pilot program? | To provide AI ethics training for researchers | To create a shared computing and data infrastructure for AI research | To establish a national AI research agenda | To develop standards for responsible AI development | To fund the creation of new AI startups | B | 5.2(a)(i) |
| 37 | Which of these is not listed as an example of an AI-related risk to critical infrastructure? | Making systems vulnerable to physical attacks | Making systems more vulnerable to cyberattacks | Causing critical infrastructure to become fully autonomous | Increasing the likelihood of critical failures | Both A and C | C | 4.3(a)(i) |
| 38 | The order requires agencies to report what to the Director of OMB annually? | AI system training data sources | AI project budgets and expenditures | Diversity statistics of AI teams | AI use cases | Number of AI patents filed | D | 10.1(e) |
| 39 | What entity or entities are directed to create specifications for nucleic acid synthesis procurement screening? | Universities and research institutions | Secretary of Commerce | Synthetic biology companies | Independent Regulatory Agencies | Department of Homeland Security | B | 4.4(b)(iv) |

| # | Question | A | B | C | D | E | Ans | Ref |
|---|---|---|---|---|---|---|---|---|
| 40 | Guidance from the Secretary of Labor is meant to protect workers from what type of harm? | AI-related collection and use of data | Not being fairly compensated for hours worked | Unfair or biased evaluation of job applicants | Job displacement risks | All of the above | E | 6(b)(iii) |
| 41 | Which entity is encouraged to prioritize AI projects for Technology Modernization Fund investments? | General Services Administration | Office of Personnel Management | Department of Veterans Affairs | National Science Foundation | None of the above | E | 10.1(g) |
| 42 | What is the stated goal of the government AI testbeds to be developed under this order? | To showcase innovative government AI applications | To train government personnel on AI technologies | To support development of safe and secure AI systems | To provide a platform for public input on AI policies | To enable agencies to test AI solutions pre-procurement | C | 4.1(a)(ii)(B) |
| 43 | The order directs the Secretary of State to consider what action related to AI talent? | Establishing AI degree scholarship programs for foreign students | Developing a special AI talent visa category | Opening international branch campuses of US AI institutes | Creating an AI expert exchange program with allied nations | Adding AI skills to Exchange Visitor Skills List | E | 5.1(b)(iii) |
| 44 | Which of these is not a required element of the plan for AI use in government benefits programs administered at the State of local level? | Effect on access for eligible beneficiaries | Replacing agency staff with automated systems | Processes for human review of AI decisions | Analysis of AI impact on equal access and outcomes | Provide notice to beneficiaries on the use of AI | B | 7.2(b)(i) |
| 45 | The Global AI Research Agenda developed under this order will do what? | Address labor-market implications across international contexts | Propose principles for workforce training data rights and privacy | Create frameworks to govern international AI research collaboration | Propose principles for mitigating environmental harms of AI hardware | Create a framework for international for AI crisis response | A | 11(c)(ii)(B) |
| 46 | Which of these is listed as a potential topic for global AI standards development? | AI transparency and explainability | Best practices for AI training data management | Hardware specifications for AI computing infrastructure | Governance structures for international AI regulatory bodies | Methods for assigning liability in AI accidents | B | 11(b)(i)(B) |
| 47 | The order requires companies developing dual-use foundation models to report red-team testing on what AI threats? | Lowering barrier for biological weapon development | Discovery of software vulnerabilities | Possibility of self-replication | All of the above | Only B and C | D | 4.2(a)(i)(C) |
| 48 | Which entity is directed to issue guidance on the use of AI in the education sector? | Department of Education | National Science Foundation | Department of Labor | White House Office of Science and Technology Policy | Department of Health and Human Services | A | 8(d) |
| 49 | The order directs the development of AI training resources for government personnel in what area? | Moral and religious implications of AI development | Policy, legal, and oversight issues related to AI | Technical maintenance and troubleshooting of AI systems | Cultural competence for equitable AI deployment | Personal cybersecurity and social media best practices | B | 10.2(g) |
| 50 | What is the purpose of the Artificial Intelligence Governance Board to be established in agencies? | To approve AI budgets and contracts | To coordinate AI activities across the agency | To provide ethical oversight of AI systems | To handle public communications about agency AI work | To lead AI research and development projects | B | 10.1(b)(iii) |
| 51 | Which entity is tasked with establishing a framework for assessing nucleic acid screening methods? | Office of Science and Technology Policy | National Institutes of Health | National Science Advisory Board for Biosecurity | National Center for Biotechnology Information | Department of Homeland Security | A | 4.4(b)(i) |
| 52 | The order directs the US Patent and Trademark Office to issue guidance on what AI-related topic? | Qualifying AI systems for expedited patent processing | Analyzing the role of AI in inventorship | Patent application requirements for AI hardware components | Copyright protection for AI-generated creative works | Trademark registration for AI company branding | B | 5.2(c)(i) |
| 53 | What AI technical skills are suggested to be important among law inforcement professionals? | Physical security, energy use signatures, AI supply chain | Data privacy and ethical data practices | Cybersecurity and red-teaming | Machine learning, data privacy, and infrastructure | Social service provision and community outreach | D | 7.1(c) |
| 54 | The order directs the Administrator of General Services to take what action related to AI procurement? | Create a certification program for AI vendors | Establish a government-wide AI solutions catalog | Prohibit agencies from purchasing foreign-developed AI | Mandate open-source licensing for procured AI systems | Facilitate access to government-wide acquisition solutions | E | 10.1(h) |
| 55 | What is the stated purpose of the government-wide method for AI tracking and assessment directed in the order? | To track the capacity to adopt AI, comply with Federal policy, and manage risks | To determine hiring needs and training needs for specific AI skills | To assess regulatory progress in Sector Risk Management Agencies | To inform development of AI career paths | To evaluate agencies on their rate of AI adoption | A | 10.1(c) |
| 56 | The order requires all government data used in AI training to have what property? | Obtained voluntarily from data subjects | Available in machine-readable formats | Stripped of all personally identifiable information | Reviewed for potential security risks | Licensed as intellectual property of the government | D | 4.7(b) |
| 57 | Which of these is not a focus area for standards and tools related to generative AI? | Preventing generation of child sexual abuse material | Preventing hallucinations | Techniques for detecting synthetic content | Tracking the provenance of synthetic content | Watermarking synthetic content | B | 4.5(a) |
| 58 | The order directs agencies to make what change related to AI talent hiring? | Require AI to be a mandatory hiring preference | Eliminate minimum education requirements for AI roles | Expand existing pay incentives to hire AI talent | Create internships positions for AI | Run recruiting fairs at top colleges nationwide | C | 10.2(e) |
| 59 | Which entity is tasked with establishing a position description library for data scientists? | Chief AI Officers Council | Chief Data Officers Council | Chief Information Officers Council | Chief Human Capital Officers Council | Chief Acquisition Officers Council | B | 10.2(f) |
| 60 | The order requires the Director of OMB to take what action related to AI in government? | Establish an AI oversight and accountability office | Develop an AI risk assessment framework for agencies | Issue requests for necessary legislation related to AI | Convene an interagency council to coordinate AI activities | Require agencies to appoint Chief AI Ethics Officers | D | 10.1(a) |
| 61 | Which is part of expanded agency AI use case reporting under the order? | Description of the AI system's purpose and function | Reporting on risks and mitigation measures | Estimated budget and personnel resources required | Performance metrics and evaluation results | Whether system is deployed or under development | B | 10.1(e) |
| 62 | In granting authorizations to operate IT systems, agencies are to prioritize what type of systems? | Systems using cloud computing infrastructure | Systems targeted at AI assurance | Systems developed by small businesses | Generative AI and emerging technologies | Systems using open-source AI components | D | 10.1(f)(ii) |

| # | Question | A | B | C | D | E | Ans | Ref |
|---|---|---|---|---|---|---|---|---|
| 63 | The Consumer Financial Protection Bureau is encouraged to use its authority to prevent what type of harm? | AI deep fakes and unauthorized sexual images | Bias in real estate appraisal and credit rating | Automation-spurred job displacement | Racial bias in AI-powered facial recognition technology | Use of AI to deliver personalized advertising or political propaganda | B | 7.3(b) |
| 64 | The order directs the Department of Commerce to engage with what type of entity on nucleic acid screening? | Biodefense and biosecurity agencies | Pharmaceutical and biotechnology companies | Academic and research institutions | Genetic sequencing and synthesis laboratories | All industry stakeholders deemed relevant | E | 4.4(b)(ii) |
| 65 | What concern does the order raise about data procured from commercial data vendors? | Managing risk of personally identifiable information in the data | Contractual guaranties to ensure regular updates to data sets | Ability to track the provenance of data back to its original sources | Purchasing data whose original source is the federal government | Purchasing commercial data obtained without documented user consent | A | 10.1(b)(viii) |
| 66 | Which of the following best describes the order's approach to AI governance? | Prohibition of AI systems deemed high-risk | Voluntary self-regulation by AI companies | Coordinated, risk-based, government-wide approach | Deregulation of AI development and use | Encouraging government-sponsored AI research and development | C | 1, 2, 4, 7, 8, 10 |
| 67 | The order aims to position the US as a leader in AI by focusing on which key areas? | Innovation and competition | Equity and civil rights | Supporting workers | Safety and security | All of the above | E | 1, 2, 5, 6, 8, 11 |
| 68 | Which of these is a way the order aims to mitigate AI risks related to privacy? | Prohibiting collection of personal data for AI purposes | Mandating use of privacy-preserving technologies | Supporting efforts to obtain broad individual consent for AI data use | Ensuring the collection, use, and retention of personal data is lawful | Requiring disclosure of AI training data sources | D | 2(f), 9(a) |
| 69 | The order emphasizes the importance of protecting what group from AI-related harms? | Government agencies and employees | AI researchers and developers | Private sector companies investing in AI | General public and consumers | Foreign nationals and international organizations | D | 1, 2(c), 2(d), 2(e), 7, 8 |
| 70 | What is a key focus of the order's guidance on AI in law enforcement and criminal justice? | AI-based predictive policing methods | AI-assisted judicial decision-making | Mitigating risks of bias and discrimination | Use of algorithms to reduce criminal recidivism | Defining liability for the use of AI in the justice system | C | 7.1 |
| 71 | The order directs agencies to establish what type of role related to AI governance? | AI Ombudsman | AI Ethics Compliance Officer | AI Public Affairs Liaison | AI Workforce Training Coordinator | Chief AI Officer | E | 10.1(b)(i) |
| 72 | Which of these best describes the order's approach to international AI cooperation? | Engaging allies to develop common AI principles and standards | Restricting AI technology transfer to strategic rivals | Teaming with allies on cutting-edge AI research and development | Encouraging bilateral AI development partnerships | Establishing treaties to prevent an AI arms race with strategic rivals | A | 11 |
| 73 | The order directs the development of an "AI toolkit" for which sector? | Banking and finance | Criminal justice | Healthcare and biomedicine | Education | All of the above | D | 8(d) |

| # | Question | A | B | C | D | E | Answer | Ref |
|---|---|---|---|---|---|---|---|---|
| 74 | What is a key goal of the AI red-teaming guidelines to be developed under the order? | Assessing system performance and accuracy | Ensuring interoperability with legacy systems | Assess limitations, vulnerabilities, and harmful behaviors | Evaluating user experience and interface design | Optimizing computational efficiency and speed | C | 4.1(a)(ii) |
| 75 | Under this order, which entity is charged with promoting the use of privacy-enhancing technologies? | Cybersecurity and Infrastructure Security Agency | Office of Science and Technology Policy (OSTP) | National Science Foundation | United States Digital Service | National Institute of Standards and Technology (NIST) | C | 9(c)(ii) |
| 76 | Which of these is not a required element of the AI risk management framework for agencies? | Conducting public consultation and engagement | Continuously monitoring deployed AI systems | Countering algorithmic discrimination | Implementing differential privacy techniques for all data | Providing avenues for redress of adverse AI decisions | D | 10.1(b)(iv) |
| 77 | The order directs the Department of Energy to develop what type of facility? | Centralized database of critical energy infrastructure | AI facilities capable of handing classified and sensitive national security data | AI testbeds | High-performance computing clusters equipped with GPUs | Quantum computing and cryptographic systems | C | 4.1.b |
| 78 | Which entity is tasked with developing best practices for AI use in benefits administration? | Department of Health and Human Services | Social Security Administration | General Services Administration | Department of Veterans Affairs | All of the above | A | 7.2(b)(i) |
| 79 | The order encourages agencies to enable access to what type of AI system for experimentation? | Open source AI models and components | Generative AI and large language models | Facial recognition and biometric identification | High-performance computing clusters equipped with GPUs | Quantum computing and cryptographic systems | B | 10.1(f)(i) |
| 80 | Which of these is a focus of the AI education and training directed in the order? | Technical skills for AI system development and maintenance | Ethical considerations and responsible AI principles | AI project management and budgeting processes | Data labeling and annotation best practices | All of the above | B | 8(d), 10.2(g) |
| 81 | Under this order, who is charged with training 500 new AI researchers? | Secretary of Labor | Secretary of Commerce | Secretary of HHS | Secretary of Education | Secretary of Energy | E | 5.2(b) |
| 82 | Which entity is directed to develop an AI incident response playbook for critical infrastructure? | Cybersecurity and Infrastructure Security Agency | National Telecommunications and Information Administration | Department of Homeland Security | Department of Energy | National Security Agency | C | 4.3(a)(v) |
| 83 | The order tasks which entity with developing a global AI research agenda? | US Agency for International Development | National Science Foundation | United States Digital Service | Presidential Innovation Fellows | US National Laboratories | A | 11(c)(ii) |
| 84 | Under this order, Federal acquisition guidance should be modified to do what? | To enable agencies to purchase the most advanced AI technologies | To ensure compatibility of procured AI with legacy systems | To reduce the risks posed by synthetic content | To centralize and streamline AI acquisition processes | To mandate domestic sourcing of AI systems and components | C | 4.5.d |
| 85 | Which of these is a key consideration in regulating AI-enabled healthcare technology ? | Implications for intellectual property and patents | Potential to exacerbate health disparities and inequities | Compliance with existing rules for human subjects research | Equipping medical devices for real-time monitoring for safety issues | Pre-market assessment and post-market oversight of algorithmic performance | E | 8(b)(ii) |

| # | Question | A | B | C | D | E | Answer | Ref |
|---|---|---|---|---|---|---|---|---|
| 86 | The order directs NIST to develop guidelines to evaluate what type of AI technology? | User acceptance testing | Computational efficiency optimization | Adversarial attack resistance | Differential privacy preservation | Model interpretability and explainability | D | 9(b) |
| 87 | Which entity is encouraged to develop guidance on AI and disability rights issues? | National Council on Disability | Architectural and Transportation Barriers Compliance Board | Office of Disability Employment Policy | Administration for Community Living | President's Committee for People with Intellectual Disabilities | B | 7.3(d) |
| 88 | What recommendation does this order seek from the Secretary of Defense? | How AI can be used to expand US economic spheres of influence | How to prevent rival nations from collecting biometric data on US persons | How DoD may hire and retain noncitizen experts in AI | Strategies for working with NATO allies in mutual defense | How to prevent AI from being used for cyber attacks | C | 10.2(h) |
| 89 | Which of these is a goal of the AI safety program for healthcare directed by the order? | To enable faster regulatory approval of AI medical devices | To establish liability protection for healthcare AI companies | To develop best practices for preventing harm from medical AI | To mandate use of government-approved healthcare AI systems | To increase funding for AI-driven precision medicine initiatives | C | 8(b)(iv)(B) |
| 90 | What is a core aim of US engagement on international AI standards development? | To ensure bilateral access to foreign markets | To enhance international trade and economic growth | To promote information sharing | To address security contexts beyond United States borders | To address AI's labor-market implications across international contexts | C | 11(b)(iii) |
| 91 | The order directs the Secretary of Commerce to define reportable AI systems based on what? | Level of funding and research expenditures | Nationality and affiliations of development team | Scale of computing power used in training | Sector and intended use cases of the system | Potential societal risks and impact | C | 4.2(b) |
| 92 | Which entity is tasked with increasing public access to government data for AI purposes? | Office of Science and Technology Policy | Interagency Council on Statistical Policy | Chief Information Officer Council | Chief Data Officers Council | Federal Privacy Council | D | 4.7(a) |
| 93 | The order emphasizes the need to protect Americans against fraud, bias, and discrimination in which key sectors? | Financial services | Law | Transportation | Healthcare | All of the above | E | 8(b), 8(d), 7.2(b)(ii) |
| 94 | Which entity is reponsible for laying the groundwork for an AI Bill of Rights? | Federal Trade Commission | Department of Justice Civil Rights Division | Office of Science and Technology Policy | National Institute of Standards and Technology | Consumer Product Safety Commission | C | 2(d), 10.1(b)(iv) |
| 95 | The order directs agencies to enhance access to what type of data for AI development? | Personal data on individual consumers | Cyber attack data | Government-funded research and scientific data | Private sector proprietary and commercial data | Data related to public infrastructure and facilities | C | 4.7, 5.2(a)(i), 11(c)(ii)(B) |
| 96 | What is the primary focus of the AI safety program directed for the healthcare sector? | Protecting patient privacy and data confidentiality | Sharing data on patient harms for research purposes | Mitigating bias and ensuring equitable treatment | Identifying and capturing incidents of patient harm deriving from AI | All of the above | D | 8(b)(i)(A) (other elements included but this is primary) |
| 97 | Which of these is not a factor agencies should consider in evaluating AI system risks? | Robustness to unexpected or unintended inputs | Potential for malicious misuse or repurposing | Comparison of AI to human error rates | Security vulnerabilities and attack surfaces | Implications for individual rights and liberties | C | 4 |

| 98 | The order highlights the importance of public participation and engagement in agency AI activities through what means? | Providing explanations of decisions made by AI algorithms | Enabling individual consent for AI data collection and use | Soliciting public input and feedback on AI policies | Publicly disclosing AI system design and training data | All of the above | C | 4.6(a), 5.2(g)(iii), 10.1(b)(iv) |
| --- | --- | --- | --- | --- | --- | --- | --- | --- |
| 99 | Under this order, what approach is taken to enhance the availability of US-manufactured semiconductors for AI? | Increasing the availability of resources to startups and small businesses | Establishing a major grant program for new chip manufacturing facilities in the US | Increasing grants for research into new chip designs | Placing tariffs on imported graphics processing units (GPUs) | Using the Defense Production Act to grow production of domestic chip manufacture | A | 5.3.b |
| 100 | Which entity is directed to issue guidance on AI and trademark policy? | National Institute of Standards and Technology | US Copyright Office | US Patent and Trademark Office | Federal Trade Commission | Department of Justice Antitrust Division | C | 5.2(c)(ii) |

# Information Extraction Tasks

7. List all Federal Government Organizations (Departments, Agencies, Administrations, Centers, Institutes, Services, and Offices) mentioned in Executive Order 14110 (for example, National Institute of Standards and Technology (NIST), Department of Homeland Security, Office of Management and Budget). Include the organization if individual affiliated with the organization is mentioned (for example, Secretary, Director, Chief AI Officer). Do not include boards, councils, groups, programs, or committees.

Answer:

1. Advanced Research Projects Agency-Infrastructure (ARPA-I)
2. Consumer Financial Protection Bureau
3. Cybersecurity and Infrastructure Security Agency
4. Department of Agriculture
5. Department of Commerce
6. Department of Defense
7. Department of Education
8. Department of Energy
9. Department of Health and Human Services (HHS)
10. Department of Homeland Security
11. Department of Housing and Urban Development
12. Department of Justice
13. Department of Labor
14. Department of State
15. Department of the Treasury
16. Department of Transportation
17. Department of Veterans Affairs
18. Executive Office of the President
19. Federal Bureau of Investigation
20. Federal Communications Commission
21. Federal Energy Regulatory Commission
22. Federal Housing Finance Agency
23. Federal Trade Commission
24. General Services Administration
25. National Aeronautics and Space Administration
26. National Institute of Standards and Technology (NIST)
27. National Intellectual Property Rights Coordination Center
28. National Institutes of Health

29. National Science Foundation (NSF)
  30. National Telecommunications and Information Administration
  31. Office of Management and Budget (OMB)
  32. Office of Personnel Management (OPM)
  33. Office of Science and Technology Policy (OSTP)
  34. Office of the Director of National Intelligence
  35. Office of the National Cyber Director
  36. Office of Pandemic Preparedness and Response Policy
  37. Office of the Vice President
  38. Small Business Administration
  39. United States Agency for International Development
  40. United States Copyright Office of the Library of Congress
  41. United States Customs and Border Protection
  42. United States Digital Service
  43. United States Patent and Trademark Office (USPTO)

2. List all actions and responsibilities in Executive Order 14110 that are explicitly assigned to the Department of Health and Human Services (HHS) or its subcomponents (such as CDC, CMS, and FDA). Include actions assigned to the Secretary of HHS or other HHS personnel. Give a fully-numbered reference for each identified action (for example, 4.1, 6.b.i, 4.2.a.i.B, or 5.3.b.iii.D).

Answer:

| Reference | Action |
| --- | --- |
| 4.4.b.i | Consult with OSTP on establishing a framework to encourage providers of synthetic nucleic acid sequences to implement comprehensive, scalable, and verifiable synthetic nucleic acid procurement screening mechanisms, including standards and recommended incentives. |
| 4.4.b.ii | Consult with NIST on initiating an effort to engage with industry and relevant stakeholders to develop and refine for specifications and best practices for effective nucleic acid synthesis procurement screening. |
| 5.2.e | Prioritize grantmaking and other awards, as well as undertake related efforts, to support responsible AI development and use. |
| 7.2.b.i | Publish a plan, informed by the guidance issued pursuant to section 10.1.b of this order, addressing the use of automated or algorithmic systems in the implementation by States and localities of public benefits and services administered by the Secretary. |
| 8.b.i | Establish an HHS AI Task Force that shall, within 365 days of its creation, develop a strategic plan that includes policies and frameworks — possibly including regulatory action, as appropriate — on responsible deployment and use of AI and AI-enabled technologies in the health and human services sector. |
| 8.b.ii | Develop a strategy to determine whether AI-enabled technologies in the health and human services sector maintain appropriate levels of quality |
| 8.b.iii | Consider appropriate actions to advance the prompt understanding of, and compliance with, Federal nondiscrimination laws by health and human services providers that receive Federal financial assistance. |
| 8.b.iv | Establish an AI safety program that, in partnership with voluntary federally listed Patient Safety Organizations. |
| 8.b.v | Develop a strategy for regulating the use of AI or AI-enabled tools in drug-development processes. |

3. Create a list of actions in Executive Order 14110 that must be completed within 6 months of the issuance of the order. Output the result as a three-column table giving (1) the item number (for example, 4.5.a, 7.2.b.ii, 8.b.iii, etc.), (2) the specified deadline (e.g., 45 days, 60 days, etc.), and (3) the action to be performed.

Answer:

| Reference | Deadline | Action |
| --- | --- | --- |
| 4.2.a | 90 days | Secretary of Commerce shall require companies developing or demonstrating intent to develop dual-use foundation models and companies acquiring/developing/possessing potential large-scale computing clusters to provide certain information to the Federal Government. |
| 4.2.c | 90 days | Secretary of Commerce shall propose regulations requiring U.S. IaaS providers to submit reports on foreign persons conducting training runs of large AI models. |
| 4.2.d | 180 days | Secretary of Commerce shall propose regulations requiring U.S. IaaS providers to ensure foreign resellers verify identity of foreign persons obtaining IaaS accounts. |

| ID | Deadline | Description |
|---|---|---|
| 4.3.a.i | 90 days | Heads of agencies with regulatory authority over critical infrastructure and heads of Sector Risk Management Agencies shall evaluate and assess potential risks of AI in critical infrastructure. |
| 4.3.a.ii | 150 days | Secretary of the Treasury shall issue public report on best practices for financial institutions to manage AI-specific cybersecurity risks. |
| 4.3.a.iii | 180 days | Secretary of Homeland Security shall incorporate as appropriate the AI Risk Management Framework, NIST AI 100-1, as well as other appropriate security guidance, into relevant safety and security guidelines for use by critical infrastructure owners and operators. |
| 4.3.b.ii | 180 days | Secretary of Defense and Secretary of Homeland Security shall develop plans and conduct operational pilot projects to identify, develop, test, evaluate and deploy AI for cyber defense. |
| 4.4.a.i | 180 days | Secretary of Homeland Security shall evaluate potential for AI misuse to enable CBRN threats and consider benefits of AI to counter threats. |
| 4.4.a.ii | 120 days | Secretary of Defense shall contract with National Academies for a study on biosecurity risks. |
| 4.4.b.i | 180 days | Director of OSTP shall establish framework to encourage synthetic nucleic acid providers to implement screening mechanisms. |
| 4.4.b.ii | 180 days | Secretary of Commerce shall initiate effort to engage with industry and stakeholders to develop specifications, best practices, implementation guides, and assessment mechanisms for nucleic acid synthesis screening. |
| 5.1.a | 90 days | Secretary of State and Secretary of Homeland Security shall take steps to streamline visa processing and ensure visa appointment availability for noncitizens working on AI. |
| 5.1.b | 120 days | Secretary of State shall consider rulemaking, publishing updates, and implementing domestic visa renewal program for Exchange Visitor Skills List. |
| 5.1.c | 180 days | Secretary of State shall consider rulemaking to expand domestic visa renewal eligibility and establish program to attract overseas AI talent. |
| 5.1.d | 180 days | Secretary of Homeland Security shall review and initiate policy changes to clarify/modernize immigration pathways for AI experts. |
| 5.1.e | 45 days | Secretary of Labor shall publish RFI on AI/STEM occupations with insufficient U.S. workers. |
| 5.1.g | 120 days | Secretary of Homeland Security shall develop and publish resources to attract and retain AI experts. |
| 5.2.a.i | 90 days | Director of NSF shall launch a pilot program implementing the National AI Research Resource (NAIRR). |
| 5.2.a.ii | 150 days | Director of NSF shall fund and launch at least one NSF Regional Innovation Engine prioritizing AI-related work. |
| 5.2.b | 120 days | Secretary of Energy shall establish a pilot program to enhance existing successful training programs for scientists. |
| 5.2.c.i | 120 days | USPTO Director shall publish guidance on inventorship and AI in inventive process. |
| 5.2.d | 180 days | Director of the National Intellectual Property Rights Coordination Center shall develop a training, analysis, and evaluation program to mitigate AI-related IP risks. |
| 5.2.g | 180 days | Secretary of Energy shall take multiple actions related to electric power and clean energy. |
| 5.2.h | 180 days | President's Council of Advisors on Science and Technology shall submit to the President a report on the potential role of AI in research aimed at tackling major societal and global challenges. |
| 6.a.i | 180 days | Chairman of Council of Economic Advisers shall submit report to President on AI labor market effects. |
| 6.a.ii | 180 days | Secretary of Labor shall submit report to President analyzing agency abilities to support workers displaced by AI. |
| 6.b.i | 180 days | Secretary of Labor shall publish principles for employers to mitigate AI harms to employees. |
| 7.1.a.ii | 90 days | Assistant Attorney General in charge of the Civil Rights Division to convene a meeting of the heads of Federal civil rights offices. |
| 7.1.c.i | 180 days | Interagency working group shall identify best practices for hiring law enforcement with AI skills. |
| 7.2.b.i | 180 days | Secretary of HHS shall publish plan on use of automated systems in state/local benefits programs. |
| 7.2.b.ii | 180 days | Secretary of Agriculture shall issue guidance to state/local administrators on use of automated systems in USDA benefits programs. |
| 7.3.c | 180 days | Secretary of HUD shall issue guidance on tenant screening systems and algorithmic advertising that may violate fair housing laws. |
| 8.b.i | 90 days | Secretary of HHS shall develop a strategic plan that includes policies and frameworks on responsible deployment and use of AI and AI-enabled technologies in the health and human services sector. |
| 8.b.ii | 180 days | Secretary of HHS shall develop strategy on ensuring quality of AI in health/human services. |
| 8.b.iii | 180 days | Secretary of HHS shall consider actions to advance nondiscrimination law compliance related to AI in health/human services. |
| 8.c.i | 30 days | Secretary of Transportation shall direct the Nontraditional and Emerging Transportation Technology (NETT) Council to assess the need for information, technical assistance, and guidance regarding the use of AI in transportation. |
| 8.c.ii | 90 days | Secretary of Transportation shall direct appropriate Federal Advisory Committees of the DOT to provide advice on the safe and responsible use of AI in transportation. |
| 8.c.iii | 180 days | Secretary of Transportation shall direct the Advanced Research Projects Agency-Infrastructure (ARPA-I) to explore the transportation-related opportunities and challenges of AI. |
| 9.a.iii | 180 days | Assistant to the President for Economic Policy and the Director of OSTP shall issue an RFI to inform potential revisions to guidance to agencies on implementing the privacy provisions of the E-Government Act of 2002 (Public Law 107-347). |
| 9.c.i | 120 days | Director of NSF shall fund the creation of a Research Coordination Network (RCN) dedicated to advancing privacy research. |
| 10.1.a | 60 days | Director of OMB shall convene and chair an interagency council to coordinate the development and use of AI in agencies' programs and operations, other than the use of AI in national security systems. |
| 10.1.b | 150 days | Director of OMB shall issue guidance to agencies on AI use in government. |
| 10.1.f.ii | 90 days | Administrator of GSA shall develop framework prioritizing AI offerings in FedRAMP. |
| 10.1.f.iii | 180 days | Director of OPM shall develop guidance on workforce use of generative AI. |
| 10.1.g | 30 days | Technology Modernization Board shall consider prioritizing funding for AI projects for the Technology Modernization Fund for a period of at least 1 year. |
| 10.2.a | 45 days | Director of OSTP and Director of OMB shall identify priority areas for federal AI talent. |
| 10.2.b | 45 days | Deputy Chief of Staff for Policy shall convene AI and Technology Talent Task Force. |
| 10.2.b.i | 180 days | AI and Technology Talent Task Force shall report to President on increasing federal AI capacity. |
| 10.2.c | 45 days | U.S. Digital Service, Presidential Innovation Fellowship and other programs shall develop plans to recruit AI talent. |
| 10.2.d.i | 60 days | Director of OPM shall review need for direct-hire authority for AI roles. |
| 10.2.d.ii | 60 days | Director of OPM shall consider authorizing excepted service appointments for AI staff. |
| 10.2.d.iii | 90 days | Director of OPM shall coordinate pooled hiring action for AI talent. |
| 10.2.d.iv | 120 days | Director of OPM shall issue guidance on pay flexibilities for AI positions. |
| 10.2.d.v | 180 days | Director of OPM shall establish guidance on skills-based hiring for AI roles. |
| 10.2.d.vi | 180 days | Director of OPM shall establish interagency working group to facilitate AI hiring. |
| 10.2.d.vii | 180 days | Director of OPM shall review Executive Core Qualifications for AI literacy. |
| 10.2.d.viii | 180 days | Director of OPM shall complete review of competencies for civil engineers reflecting AI use. |
| 10.2.h | 180 days | Secretary of Defense shall submit a report to the President on hiring noncitizens and AI experts. |

4. Which specific actions in Executive Order 14110 deal with assuring AI is applied equitably, without bias, in a non-discriminatory manner? Return a list of actions, a fully-numbered reference for each identified action (for example, 4.1, 6.b.i, 4.2.a.i.B, or 5.3.b.iii.D), and the name of the agency involved with implementing the measure.

Answer:

| Reference | Action |
|---|---|
| 5.2.e.iii | Accelerate grants awarded through the National Institutes of Health Artificial Intelligence/Machine Learning Consortium to Advance Health Equity and Researcher Diversity (AIM-AHEAD) program and showcasing current AIM-AHEAD activities in underserved communities. |
| 5.2.g | Secretary of Energy shall support the goal of strengthening our Nation's resilience against climate change impacts and building an equitable clean energy economy for the future while improving environmental and social outcomes. |
| 6.b.i.B | Secretary of Labor shall publish principles and best practices for employers addressing labor standards and job quality, including issues related to the equity. |
| 7.1.a | The Attorney General shall coordinate with agencies to implement and enforce laws to address civil rights violations and discrimination related to AI. |
| 7.1.b.i | The Attorney General shall report to the President on the use of AI in the criminal justice system. |
| 7.1.b.ii | The Attorney General shall identify areas where AI can enhance law enforcement efficiency and accuracy, consistent with protections for privacy, civil rights, and civil liberties. |

| 7.1.c.iii | The Attorney General shall review the work conducted pursuant to section 2.b of Executive Order 14074 and, if appropriate, reassess the existing capacity to investigate law enforcement deprivation of rights under color of law resulting from the use of AI. |
|---|---|
| 7.2.a | Agencies shall use civil rights and civil liberties offices to prevent and address unlawful discrimination and harms from AI in Federal programs and benefits administration, |
| 7.2.b.i | The Secretary of HHS shall publish a plan addressing the use of AI systems in public benefits to promote equitable access, notice, evaluation to detect unjust denials, and analysis of whether systems achieve equitable outcomes. |
| 7.2.b.ii | The Secretary of Agriculture shall issue guidance to benefits administrators on using AI systems in an equitable manner. |
| 7.3.a | The Secretary of Labor shall publish guidance for Federal contractors on non-discrimination in hiring involving AI. |
| 7.3.b | FHFA and CFP should use their authority to assure regulated entities evaluate their underwriting and automated collateral-valuation and appraisal processes in ways that minimize bias. |
| 7.3.c | The Secretary of HUD and Director of CFPB shall issue guidance on preventing algorithmic discrimination in housing transactions. |
| 7.3.d | The Architectural and Transportation Barriers Compliance Board should help ensure that people with disabilities benefit from AI's promise while being protected from its risks, including unequal treatment from the use of biometric data. |
| 8.a | Independent regulatory agencies are encouraged, as they deem appropriate, to consider using their full range of authorities to protect American consumers from fraud, discrimination, and threats to privacy and to address other risks that may arise from the use of AI. |
| 8.b.i.C | The HHS AI Task Force strategic plan shall address incorporation of equity principles in AI-enabled technologies used in health and human services. |
| 8.b.iii | Secretary of HHS shall consider appropriate actions to advance the prompt understanding of, and compliance with, Federal nondiscrimination laws by health and human services providers that receive Federal financial assistance |
| 8.b.iv.A | Secretary of HHS shall establish a clinical safety program that tracks harms caused by AI, including through bias or discrimination |
| 8.d | The Secretary of Education shall develop policies on non-discriminatory uses of AI in education, including impacts on vulnerable and underserved communities. |
| 10.2.b.ii | AI and Technology Talent Task Force shall identify and circulate best practices for agencies to attract, hire, retain, train, and empower AI talent, including diversity, inclusion, and accessibility best practices. |
| 10.1.b.iv | OMB shall issue guidance specifying required risk management practices for AI that impacts rights and safety, including assessing and mitigating disparate impacts and algorithmic discrimination. |
| 10.1.b.viii.B | Director of OMB shall issue recommendations on testing and safeguards against discriminatory, misleading, inflammatory, unsafe, or deceptive outputs for generative AI. |

5. Identify actions in Executive Order 14110 that apply to all US Federal Government agencies. Return a list of actions, a fully-numbered reference for each identified action (for example, 4.1, 6.b.i, 4.2.a.i.B, or 5.3.b.iii.D), and the name of the agency involved with implementing the measure.

Answer:

| Reference | Action |
|---|---|
| 2.f | Agencies shall use available policy and technical tools, including privacy-enhancing technologies (PETs) where appropriate, protect privacy and to combat the broader legal and societal risks — including the chilling of First Amendment rights — that result from the improper collection and use of people's data |
| 4.7.b | Agencies shall conduct a security review of all data assets in the comprehensive data inventory required under 44 U.S.C. 3511.a.1) and (2.B) and shall take steps, as appropriate and consistent with applicable law, to address the highest-priority potential security risks that releasing that data could raise with respect to CBRN weapons, such as the ways in which that data could be used to train AI systems |
| 6.b.ii | Heads of agencies shall consider, in consultation with the Secretary of Labor, encouraging the adoption of these guidelines in their programs to the extent appropriate for each program and consistent with applicable law. |
| 7.2.a | Agencies shall use their respective civil rights and civil liberties offices and authorities — as appropriate and consistent with applicable law — to prevent and address unlawful discrimination and other harms that result from uses of AI in Federal Government programs and benefits administration |
| 10.1.f.i | Agencies are discouraged from imposing broad general bans or blocks on agency use of generative AI. |
| 10.1.g | Agencies are encouraged to submit to the Technology Modernization Fund project funding proposals that include AI — and particularly generative AI — in service of mission delivery. |
| 10.2.e | Agencies shall use all appropriate hiring authorities, including Schedule A(r) excepted service hiring and direct-hire authority, as applicable and appropriate, to hire AI talent and AI-enabling talent rapidly |

| 10.2.g | The head of each agency shall implement — or increase the availability and use of — AI training and familiarization programs for employees, managers, and leadership in technology as well as relevant policy, managerial, procurement, regulatory, ethical, governance, and legal fields |
|---|---|

6. List all new reports, guidance, regulations, or similar documents that are to be developed in response to Executive Order 14110. Return a list with a fully-numbered reference for each identified document (for example, 4.1, 6.b.i, 4.2.a.i.B, or 5.3.b.iii.D).

Answer:

| Reference | Document | Author |
|---|---|---|
| 4.1.a.i | Guidelines and best practices, with the aim of promoting consensus industry standards, for developing and deploying safe, secure, and trustworthy AI systems | NIST |
| 4.1.a.ii | Guidelines for red teaming | NIST |
| 4.2.a.i | Provide the Federal Government with information, reports, or records regarding ongoing or planned activities related to training, developing, producing, owning, securing, and testing dual-use foundation models. | Companies developing or demonstrating an intent to develop potential dual-use foundation models |
| 4.2.b | Definition of technical conditions for models and computing clusters that would be subject to the reporting requirements of subsection 4.2.a. | Commerce |
| 4.2.c.i | Regulations that require United States IaaS Providers to submit a report to the Secretary of Commerce when a foreign person transacts with that United States IaaS Provider to train a large AI model. | Commerce |
| 4.2.c.ii | Report to the Secretary of Commerce detailing each instance in which a foreign person uses the United States IaaS Product to conduct a training run described in subsection 4.2.c.i. | Foreign resellers of US IaaS products |
| 4.2.d | Regulations that require United States IaaS Providers to ensure that foreign resellers of United States IaaS Products verify the identity of any foreign person that obtains an IaaS account (account) from the foreign reseller. | Commerce |
| 4.3.a.ii | Public report on best practices for financial institutions to manage AI-specific cybersecurity risks. | Treasury |
| 4.3.a.iii | Safety and security guidelines for use by critical infrastructure owners and operators, incorporating the AI Risk Management Framework, NIST AI 100-1. | Secretary of Homeland Security |
| 4.3.b.iii | Report to the Assistant to the President for National Security Affairs on the results of actions taken pursuant to the plans and operational pilot projects required by subsection 4.3.b.ii. | Defense and the Homeland Security |
| 4.4.a.i.B | Report on the potential for AI to be misused to enable the development or production of CBRN threats. | Homeland Security |
| 4.4.b.i | Framework, incorporating existing United States Government guidance, to encourage providers of synthetic nucleic acid sequences to implement comprehensive, scalable, and verifiable synthetic nucleic acid procurement screening mechanisms. | Homeland Security |
| 4.4.b.ii | Specifications and best practices for effective nucleic acid synthesis procurement screening. | Commerce |
| 4.4.b.iv | Framework to conduct structured evaluation and stress testing of nucleic acid synthesis procurement screening and annual report on activities conducted pursuant to subsection 4.4.b.iv.A including recommendations, on how to strengthen nucleic acid synthesis procurement screening. | Homeland Security |
| 4.5.a | Report identifying the existing standards, tools, methods, and practices for detecting and authenticating synthetic content. | Commerce |
| 4.5.b | Guidance regarding the existing tools and practices for digital content authentication and synthetic content detection measures. | Commerce |
| 4.5.c | Guidance to agencies for labeling and authenticating digital content that they produce or publish. | OMB |
| 4.6.b | Report on potential benefits, risks, and implications of dual-use foundation models for which the model weights are widely available. | Commerce |
| 4.7.a | Initial guidelines for performing security reviews, including reviews to identify and manage the potential security risks of releasing Federal data that could aid in the development of CBRN weapons as well as the development of autonomous offensive cyber capabilities. | Chief Data Officer Council |
| 4.8 | National Security Memorandum on AI. | Assistant to the President for National Security Affairs and the Assistant to the President and Deputy Chief of Staff for Policy |
| 5.1.e | Request for information identifying AI and other STEM-related occupations for which there is an insufficient number of United States workers. | Labor |
| 5.1.g | Informational resources to better attract and retain experts in AI and other critical and emerging technologies. | Homeland Security |
| 5.1.g.ii | Public report with relevant data on applications, petitions, approvals, and other key indicators of how experts in AI and other critical and emerging technologies have utilized the immigration system through the end of Fiscal Year 2023. | Homeland Security |
| 5.2.a.i | Report identifying the agency resources that could be developed and integrated into a pilot program implementing the National AI Research Resource (NAIRR). | Agencies identified by NSF |
| 5.2.c.i | Guidance to USPTO patent examiners and applicants addressing inventorship and the use of AI, including generative AI, in the inventive process. | NSF |
| 5.2.c.ii | Updated guidance on patent eligibility to address innovation in AI and critical and emerging technologies. | USPTO Director |
| 5.2.c.iii | Recommendations on potential Executive actions relating to copyright and AI. | Copyright Office of the Library of Congress |
| 5.2.d.iii | Guidance and other appropriate resources to assist private sector actors with mitigating the risks of AI-related IP theft. | Homeland Security |
| 5.2.g.i | Report on the potential for AI to improve planning, permitting, investment, and operations for electric grid infrastructure and to enable the provision of electric power. | Energy |
| 5.2.h | Report on the potential role of AI, especially given recent developments in AI, in research aimed at tackling major societal and global challenges. | President's Council of Advisors on Science and Technology |
| 6.a.i | Report on labor-market effects of AI. | Council of Economic Advisers |
| 6.a.ii | Report analyzing the abilities of agencies to support workers displaced by the adoption of AI. | Labor |
| 6.b.i | Principles and best practices for employers that could be used to mitigate AI's potential harms to employees' well-being. | Labor |
| 6.b.iii | Guidance to make clear that employers that deploy AI to monitor or augment employees' work must continue to comply with protections that ensure that workers are compensated for their hours worked. | Labor |
| 7.1.b.i | Report that addresses the use of AI in the criminal justice system. | Attorney General |
| 7.2.b.i | Plan to address the use of automated or algorithmic systems in the implementation by States and localities of public benefits and services administered by the Secretary. | HHS |
| 7.2.b.ii | Guidance to State, local, Tribal, and territorial public-benefits administrators on the use of automated or algorithmic systems in implementing benefits. | Agriculture |
| 7.3.a | Guidance for Federal contractors regarding nondiscrimination in hiring involving AI and other technology-based hiring systems. | Labor |
| 7.3.c (i-ii) | Additional guidance on how Fair Housing Act applies to algorithms in real estate and for people with disabilities. | Housing and Urban Development |
| 8.b.i | Strategic plan that includes policies and frameworks — possibly including regulatory action, as appropriate — on responsible use of AI and AI-enabled technologies in the health and human services sector. | HHS |
| 8.b.ii | Strategy to determine whether AI-enabled technologies in the health and human services sector maintain appropriate levels of quality. | HHS |
| 8.d | Resources, policies, and guidance regarding AI. | Education |
| 9.a.iii | Request for Information (RFI) to inform potential revisions to guidance to agencies on implementing the privacy provisions of the E-Government Act of 2002 (Public Law 107-347). | Assistant to the President for Economic Policy and the Director of OSTP |
| 9.b | Guidelines for agencies to evaluate the efficacy of differential-privacy-guarantee protections, including for AI. | Commerce/NIST |
| 10.1.b | Guidance to agencies to strengthen the effective and appropriate use of AI, advance AI innovation, and manage risks from AI in the Federal Government. | OMB |
| 10.1.d.i | Guidelines, tools, and practices to support implementation of the minimum risk-management practices described in subsection 10.1.b.iv. | Commerce |
| 10.1.e | Instructions to agencies for the collection, reporting, and publication of agency AI use cases, pursuant to section 7225.a) of the Advancing American AI Act. | OMB |

| 10.1.f.ii | Framework for prioritizing critical and emerging technologies offerings in the Federal Risk and Authorization Management Program authorization process. | Administrator of General Services |
|---|---|---|
| 10.1.f.iii | Guidance on the use of generative AI for work by the Federal workforce. | Office of Personnel Management (OPM) |
| 10.2.b.i | Report and recommendations for further increasing AI capacity in the Federal Government. | Assistant to the President and Deputy Chief of Staff for Policy |
| 10.2.d.iv | Guidance for agency application of existing pay flexibilities or incentive pay programs for AI. | OPM |
| 10.2.d.v | Guidance and policy on skills-based, Federal Government-wide hiring of AI, data, and technology talent. | OPM |
| 10.2.f | A position-description library for data scientists (job series 1560) and a hiring guide to support agencies in hiring data scientists | Chief Data Officer Council |
| 10.2.h | Report on gaps in AI talent for national defense | Defense |
| 11.b.ii | Report on priority actions taken pursuant to the plan to advance responsible global technical standards for AI development | Commerce |
| 11.c.i | AI in Global Development Playbook that incorporates the AI Risk Management Framework's principles, guidelines, and best practices into the social, technical, economic, governance, human rights, and security conditions of contexts beyond United States borders | State and the Administrator of the United States Agency for International Development |
| 11.c.ii | Global AI Research Agenda to guide the objectives and implementation of AI-related research in contexts beyond United States borders. | Secretary of State and the Administrator of the United States Agency for International Development |
| 11.d.ii | Report to the President on priority actions to mitigate cross-border risks to critical United States infrastructure | Homeland Security |